\documentclass[conference]{IEEEtran}
\usepackage{cite}
\usepackage{graphicx}

\graphicspath{ {images/} }
\IEEEoverridecommandlockouts
\usepackage{cite}
\usepackage{amsmath,amssymb,amsfonts,bbm}
\usepackage{subfigure,bm,balance}
\usepackage[ruled,vlined]{algorithm2e}
\usepackage{graphicx}
\usepackage{textcomp}
\usepackage{xcolor}

\begin{document}

%Here goes the title

\title{SISE-PC: Semi-supervised Image Subsampling for Explainable Pathology Classification}

%Authors List

\author
{\IEEEauthorblockN{Sohini Roychowdhury}
\IEEEauthorblockA{Director, Curriculum, \\
Fourthbrain.ai, roych@uw.edu}
\vspace{-0.7cm}
\and
\IEEEauthorblockN{Kwok Sun Tang}
\IEEEauthorblockA{University of Illinois\\
kwoksun2@illinois.edu}
\vspace{-0.7cm}
\and
\IEEEauthorblockN{Mohith Ashok}
\IEEEauthorblockA{AggDirect \\
mohithashok@gmail.com}
\vspace{-0.7cm}
\and
\IEEEauthorblockN{Anoop Sanka}
\IEEEauthorblockA{Fourthbrain.ai\\
anoopsanka@gmail.com }
\vspace{-0.7cm}
}
\maketitle

%Main body starts

\begin{abstract}
Although automated pathology classification using deep learning (DL) has proved to be predictively efficient, DL methods are found to be data and compute cost intensive. In this work, we aim to reduce DL training costs by pre-training a Resnet feature extractor using SimCLR contrastive loss for latent encoding of OCT images. We propose a novel active learning framework that identifies a minimal sub-sampled dataset containing the most uncertain OCT image samples using label propagation on the SimCLR latent encodings. The pre-trained Resnet model is then fine-tuned with the labelled minimal sub-sampled data and the underlying pathological sites are visually explained. Our framework identifies upto 2\% of OCT images to be most uncertain that need prioritized specialist attention and that can fine-tune a Resnet model to achieve upto 97\% classification accuracy. The proposed method can be extended to other medical images to minimize prediction costs.
\end{abstract}

\begin {IEEEkeywords}
SimCLR, contrastive loss, Resnet, semi-supervised, label spreading
\end{IEEEkeywords}

\section{Introduction}
\label{intro}
Automated pathology classification has shown to significantly improve patient prioritization and resourcefulness of treatment procedures and patient care \cite{kagglepaper}. Although deep learning algorithms such as Resnet and InceptionV3 have been established as state-of-the-art \cite{capsulenet} for several pathology classification tasks, training these models from scratch can be expensive from the labelled data acquisition and compute resource perspectives. In this work, we present a semi-supervised image sub-sampling method that identifies a minimal sub-sampled data set that represents the most sample uncertainty in a latent feature space that is encoded using a self-supervised contrastive model \cite{simclr}. We demonstrate the classification performance of a pre-trained Resnet model that is fine tuned using only the minimal sub-sampled data for multi-class pathology classification.

\begin{figure}[ht!]
     \centering
    \includegraphics[width=0.5\textwidth]{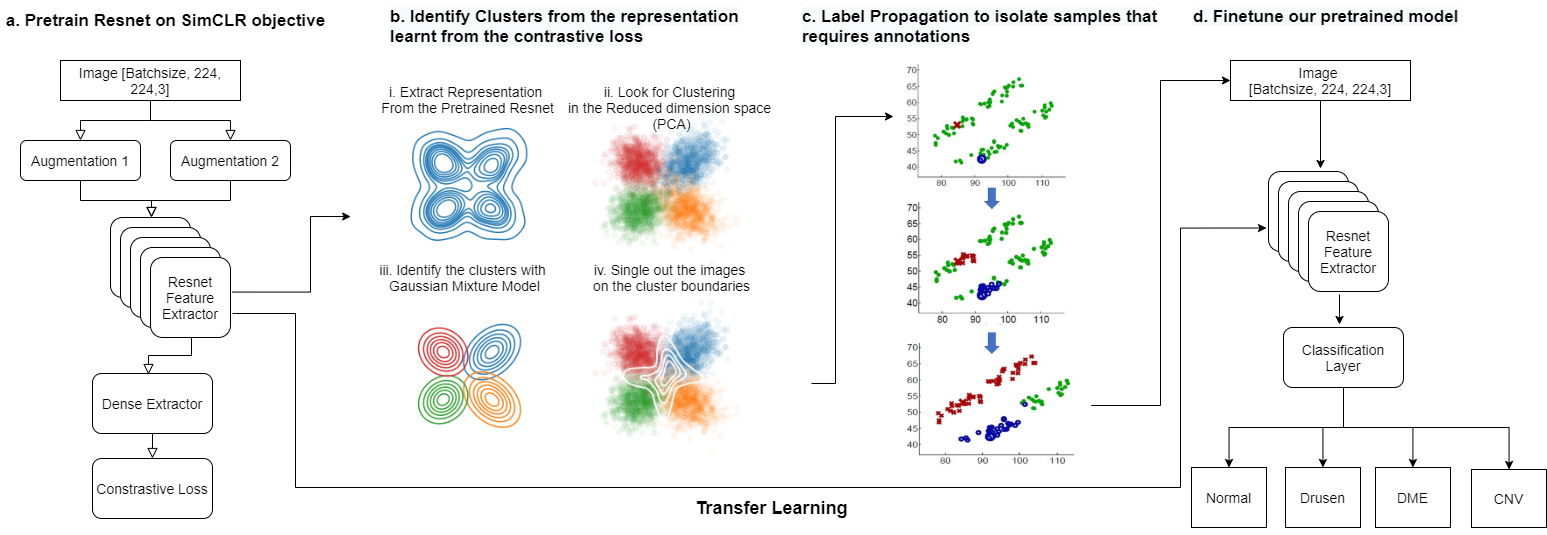}
        \caption{Overview of the proposed workflow. a) A unsupervised model SimCLR is trained on the constrastive learning objective to learn useful representation of each image. b) Dimension reduction is applied followed by unsupervised clustering and identification of cluster peripheral samples. c) Label Propagation is applied using a pre-labelled annotated image set. Images with the most uncertain labels are selected as the optimal training sample subset. d) The Resnet pretrained with SimCLR is now fine-tuned with the annotated images selected in c) for classification.}
    \label{fig:algorithm}
    \vspace{-0.4cm}
\end{figure}

%%%%%%%%%%%%% Start Here%%%%%%%%%%%%%%%%%%%%%%%%%
Optical Coherence Tomography (OCT) is a commonly performed diagnostic test designed to assist doctors in identifying retinal diseases, such as choroidal neovascularization (CNV), Diabetic macular edema (DME), and Drusen, that are the most common pathologies resulting in acquired blindness \cite{kagglepaper}. With approximately 30 million invasive OCT scans being performed each year worldwide, there is a need to identify patients with OCT images that need prioritized attention. The existing work in \cite{capsulenet} shows 92-99\% classification accuracy for OCT image classification using large annotated OCT image data sets to train a CapsuleNet model. Contrarily, in this work we propose a novel active learning framework \cite{activelearning} \cite{hao2021} to significantly reduce the training complexity by reducing the size of annotated training data set. The proposed system and steps are explained in the Fig. \ref{fig:algorithm}.

This paper makes three key contributions. First, we present a novel semi-supervised framework to identify the most uncertain set of images that are not straightforward to classify or annotate. The proposed method can identify upto 2\% of training images samples that are encoded using a self-supervised method \cite{simclr} and that are found to lie along classification decision boundaries. The proposed active learning algorithm enables prioritization of patients with such uncertain images to be seen early for treatment. Second, we present a two-step classifer training method where feature extractor layers are first pre-trained using image augmentations without labels and SimCLR \cite{simclr} framework. Next, the classifier is fine-tuning using the labelled uncertain image set for training. This process achieves 96.45\% classification test accuracy for OCT-based pathologies by training on atmost 2000 images only.
Third, we utilize the Gradcam library to analyze the regions of interest (ROIs) \cite {kagglepaper} that explain underlying pathology. We apply random image occlusion followed by feature analysis to identify primary and secondary ROIs for pathology to lie between the external limiting membrane and choroidal layers. 
\section{Data and Methods}
\label{methods}
The data set and methods used in this work are described below.
\subsection{Data: OCT Image Dataset}
In this work we apply the OCT data set in \cite{kagglepaper}, where, training data set contains images annotated for 4 classes \{CNV, DME, Drusen, Normal\} with \{26318, 8616, 11350, 37205\}, images per class, respectively. The test set has 968 images with the same 4 class labels.
\subsection{Methods and Mathematical Frameworks}
The first step to medical image classification is feature extraction while preserving the structural, contextual and textural features. The methods used to encode the OCT images to a latent vector space, and the proposed semi-supervised image sampling methods are described below \footnote{Code is available at https://github.com/anoopsanka/retinal\_oct}.

\subsubsection{Image Encoding by Self-supervised Learning}
SimCLR \cite{simclr} is a recent model proposed for unsupervised visual representation learning. A minibatch of $n$ images is sampled and applied with pairs of augmentation to produce $2n$ images. These augmented pairs are fed into a feature encoder (Resnet model) followed by a projection layer (Multilinear Perceptron) to obtain an latent vector ${\bf z}$. The training objective is to maximize the agreement of the latent vectors from the same image with augmented views (positive pair), while repulsing from other different images ($2n-1$ negatives). In SimCLR, the similarity between two views is defined using cosine similarity, i.e. $\rm{sim({\bf u,v})} = {\bf u}^T {\bf v} / ||{\bf u}|| ~ ||{\bf v}||$. Thus, the loss function for each positive pair of sample $(a, b)$ is defined to be
\begin{equation}
    \mathcal{L}_{a,b} = - \rm{log} \frac{ exp(\rm{sim({\bf z_a, z_b})}/ \tau)} { \sum_{\substack{c=1 \\ c \neq a}}^{2n} exp(\rm{sim({\bf z_a, z_c})/ \tau } )  },
\end{equation}
where $\tau$ refers to the temperature parameter. 

\begin{figure}
    \centering
    \includegraphics[width=3in, height=1.8in]{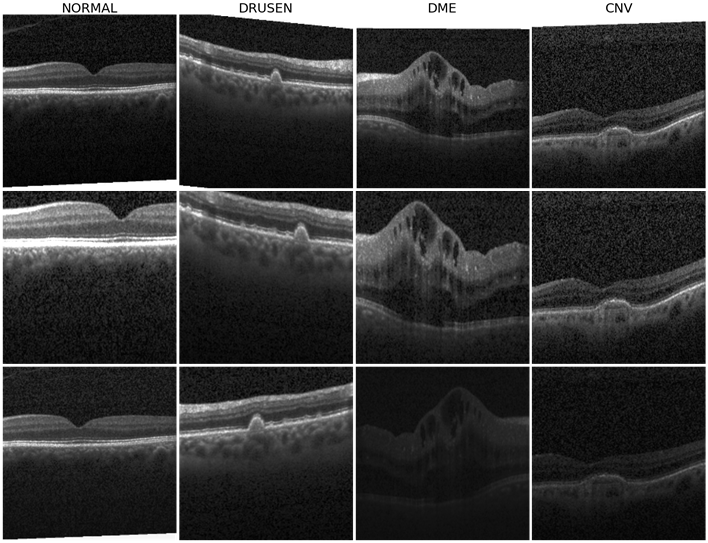}
    \caption{Row 1 shows the original image with different label class. Rows 2, 3 show the same images in Row 1 with SimCLR default augmentations.}
    \label{fig:aug_image}
    \vspace{-0.5cm}
\end{figure}

The SimCLR model is trained with batch size of 128 for 120 epochs. A batch size of 128 provides 255 negative samples per positive pair from two augmented views. We apply the LAMB optimizer since training with a large batch size may lead to instability using SGD with momentum. Similar to the original SimCLR implementation in \cite{simclr}, we use a linear warmup for the first 10 epochs and decay the learning rate with the cosine decay schedule. Here, the default augmentation strategy from \cite{simclr} is adopted. Sample augmentations are shown in Figure \ref{fig:aug_image}. The model training is stopped at epoch 75 with a contrastive accuracy of 99.77\%.

Each OCT image $I$ is resized to [224x224] followed by SimCLR self-supervised latent vector encoding process. The output of the Resnet encoder for each image of size [7x7x512] is subjected to global averaging to extract $J$, which is a [1x512] dimensional encoded vector representation for each image. We further reduce the dimensions of the latent vector representation by subjecting $J$ to PCA-based decomposition, thereby retaining 8 top principal components per image that are found to be most representative. Thus, the data set used for subsequent selective sub-sampling is represented by \{$X \in \mathcal{R}^{[n\times d]},Y \in \mathcal{R}^{[n\times1]}$\}, where $X$ depicts sample features with $d=8$ dimensions for $n=83,484$ samples, and $Y$ represents the pathological class labels. Further, cluster-specific sub-sampling methods described below.
\subsubsection{Semi-supervised Sub-sampling}
With our intention to train a classification model with minimal training labels, we use k-means to cluster $X$ into four categories/clusters with means and standard deviations, respectively, as $[\mu_k, \Sigma_k] \forall k=[1:4]$. Next, we identify uncertain samples as the ones that lie farther away from the cluster centers and are therefore more likely to lie along the classifier decision boundaries. In \eqref{GMM}, we compute the probability for a sample $x$ to belong to cluster $k$.
\begin{align}\label{GMM}
p_k(x)=\frac{exp(- \frac{1}{2}(x-\mu_k)\Sigma_k^{-1}(x-\mu_k)^T)}{\sqrt{(2\pi)^d |\Sigma|}}
\end{align}
Samples that have normalized probabilities of belonging to any cluster distribution in the uncertain range of [0.4-0.6] can be considered to lie at cluster peripheries. These samples are collected as $(X_S \subset X),  X_S \in \mathcal{R}^{[mxd]}$, where $m<n$. Other methods such as relative z-scores and Silhouette Scores per sample were also evaluated to identify uncertain samples, but the Gaussian distribution-based sample isolation proved most effective in identifying the corner cases per cluster.
Next, we use a small set of labelled data ($L=\{X_l,Y_l\}$) with atmost $n_l=80$ labelled samples as seed to apply label propagation \cite{PLL} to all unlabelled samples in $X_S$. The goal here is identification of samples that demonstrate high variances for transductive labelling. Thus, uncertain samples that acquire different class labels at different sample runs can be indicative of the uncertain space around decision boundaries. Using the semi-supervised label propagation method shown in Algorithm \ref{algo}, we identify the most uncertain minimum sub-sampled data set $(X_T \subset X_S), X_T \in \mathcal{R}^{[r\times d]}$, $r<<n$, that needs to be labelled to adequately train a deep learning classifier \cite{Resnet}.
\begin{algorithm}[ht!]
\SetAlgoLined
\KwOut{Most uncertain sample set $X_T$}
 \KwIn {$X_S$, $L=\{X_l,Y_l\}$}
 \For{$e=$ 0 \KwTo $10$}{
  Randomly select 5 samples per class from $L$\;
  \For{$x_s \in X_S$}{
    Label-spreading  with $\alpha=0.01$, kernel$=rbf$\;
    }
    \KwRet{labels $Y_S$ for $X_S$}\;
  U[:,e] $\longleftarrow Y_S$\;
 }
 $Given$: $U \in \mathcal{R}^{[m \times 10]}, X_S\in \mathcal{R}^{[m \times d]}, X_T=[.]$\;
 \For{$j=$ 0 \KwTo m}{
 $l$=unique(U[j,:])\;
 \If{length$(l)>=5$}
 {append $X_S(j,:)$ \KwTo $X_T$\;
 }
 }
 \caption{Semi-supervised Sub-sampling}\label{algo}
 \vspace{-0.2cm}
\end{algorithm}

In Algorithm \ref{algo}, label-spreading using rbf-kernel is applied with $\alpha=0.01$ implying that once a label is acquired by a sample in $X_S$ it is less likely to be modified again. Thus, we identify the samples that acquire highly variable labels across 10 label spread runs. The minimal sub-sampled data set $X_T$ is returned for labelling and classifier training thereafter.
\subsubsection{Pathology Classification and Explainability}
\begin{figure}[ht]
     \centering
    \includegraphics[width=3.1in, height=1.9in]{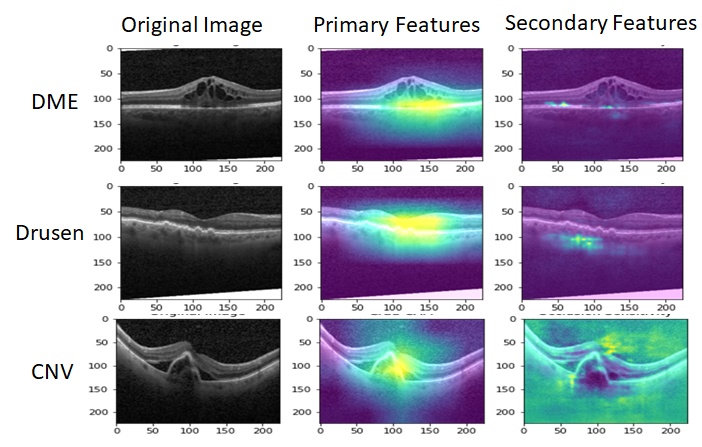}
        \caption{Examples of primary and secondary ROIs identified on test data.}
    \label{visualize}
    \vspace{-0.5cm}
\end{figure}

Once the labels for the minimal sub-sampled data set are collected, we use this minimal training data to further fine-tune the SimCLR pre-trained Resnet classifier. Next we apply the Gradcam library and tf\_explain libraries in Python \cite{gradcam} to identify primary and secondary features corresponding to ROIs that explain the underlying pathology as shown in Fig. \ref{visualize}. Here, we observe that the feature extraction layers of Resnet focus on the regions between the inner segment layer and retinal pigment epithelium layer in the OCT images to classify pathological vs. normal images. In case this ROI is occluded, the secondary regions that are analyzed for each OCT image include the choroidal regions for macular OCT images.
%% Explain medical pathologies)

\section{Experiments and Results}
\label{exp}
To enable minimal data sub-sampling followed by pathology classification, we perform three major experiments. First, we analyze the repeatability of the proposed semi-supervised sub-sampling method by analyzing across multiple runs and assessing the numbers of images detected per class. Second, we compare the classification performance of the proposed sub-sampling with random stratified sampling for Resnet classification.
Third, we analyze the robustness of features explained by the sub-sampled uncertain images.
\subsection{Analysis of Sub-sampling Repeatability} 
We repeat Algorithm \ref{algo} for 20 runs and record the numbers of images corresponding to each class category detected per run. The average and standard deviations in the number of samples per class are shown in Table \ref{tab:num}.
\begin{table}[ht!]
\centering
\caption{Analysis of sub-sampled images across 20 runs.}
\scalebox{0.99}
    {
 \begin{tabular}{|c|c|c|c|c|c|}
 \hline
Metric&Normal&Drusen&DME&CNV&Total\\ \hline
Average&711&274&261&466&1707\\ \hline
Std. dev.&78&50&26&141&251\\ \hline
\end{tabular}
 }
\label{tab:num}
\vspace{-0.4cm}
\end{table}

Here, we observe significant down sampling and relative consistency across the number of images sub-sampled across multiple runs when compared to the original sample size of 83,484 samples. Thus, the proposed method selects less than 2000 training samples, that represents about 2\% of the total sample size, to be annotated for subsequent classification. 
\subsection{Classification of sub-sampled Images} 
To analyze the importance of the novel semi-supervised sub-sampling method proposed in this work, we further fine-tune the Resnet model that was previously trained using the contrastive loss for SimCLR. For this fine-tuning, we use several sets of sub-sampled labelled data to asses the impact of transfer learning and also to identify the minimum number of samples required to train an explainable pathology classifier.

\begin{figure}[ht]
    \centering
	\subfigure[UMAP with 1\% data]
	{\includegraphics[width=0.22\textwidth]{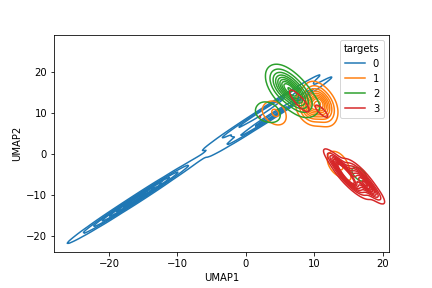}}
	\subfigure[UMAP with 98\% data]
	{\includegraphics[width=0.22\textwidth]{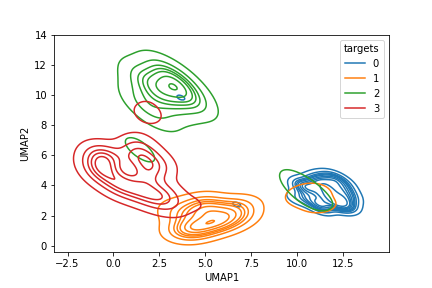}}
	\caption{UMAP representations of classified samples trained with $L$ corresponding to 1\% and 98\% stratified random samples, respectively. Separability across clusters is significantly higher with 98\% of training samples.}\label{visual}
       \vspace{-0.4cm}
\end{figure}

For this experiment, a set of labelled data ($L$) with $n_l$ samples is subjected to stratified random 80/20 train/validation split. This data set is then used to train the Resnet model layers with categorical crossentropy loss, balanced class weights and Adam Optimizer with a learning rate $10^{-5}$ . Early stopping is applied when the validation loss has not improved for more than 3 epochs. To analyze the importance of data sub-sampling over using the complete training data set, we apply stratified random sampling to isolate 1, 5, 10\% of samples per class to create labelled data $L$ that is then used to fine-tune the Resnet model. The classification performances using stratified random sub-sampling in terms of average classification accuracy, precision, recall and F1 score \cite{kagglepaper} are shown in Table \ref{res1}. 
\begin{table}[ht!]
\centering
\caption{Averaged classification performances from Stratified and Proposed Semi-supervised Sampling across several runs.}
\scalebox{0.99}
    {
 \begin{tabular}{|c|c|c|c|c|}
 \hline
Stratified Sample Size (\%)&Accuracy&Precision&Recall&F1\\\hline
1\%	&0.9101&0.9221&0.9101&0.9083\\\hline
5\% &0.9834&0.9844&0.9834&	0.9834\\\hline
10\%&0.9880&	0.9883&	0.9880&	0.9880\\\hline
100\%&0.9989&	0.9989&	0.9989&	0.9989\\\hline
\multicolumn{5}{|c|}{Proposed Semi-supervised Sampling}\\ \hline
{\bf 2\%}&{\bf 0.9645}&{\bf 0.9675}&{\bf 0.9625}&{\bf 0.9625}\\\hline
\end{tabular}
 }
\label{res1}
\vspace{-0.3cm}
\end{table}
Here, we observe that training by atmost 5\% random stratified samples, the Resnet model is significantly well trained to visually explain pathology classification. Improvements in classification performances thereafter are incremental and often not generalizable. Also, in Fig. \ref{visual} we observe the UMAP representations of classified samples with 1\% and 98\% of the data samples. We observe significant cluster separability as the training data set increases. 

Our goal is to identify the minimal training data set to achieve significant cluster separability. Thus, we analyze the performance of the proposed sub-sampling method to train the classifier in Table \ref{res1}. We observe that the proposed sub-sampled data set that isolates 2\% training images around decision boundaries achieves about 96.45\% classification accuracy. This performance is comparable to InceptionV3 based model in \cite{capsulenet} that achieved 96.1\% accuracy with complete training data set. 
Thus, the proposed sub-sampling method intuitively lies within the 5\% sampling size as learned from the random stratified sampling. Additionally, the Resnet model fine-tuned on the proposed minimal sub-sampled data achieves consistent precision, and recall performances, which indicates consistencies in the numbers of classified false positives and false negatives.

\begin{figure}[ht]
    \centering
	\subfigure[1\% Stratified Random sampling.]
	{\includegraphics[width=0.22\textwidth]{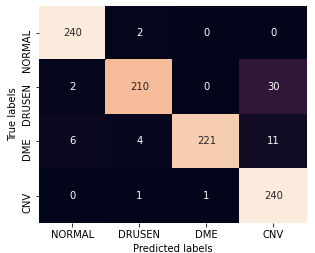}}
	\subfigure[2\% Proposed Sub-sampling.]
	{\includegraphics[width=0.22\textwidth]{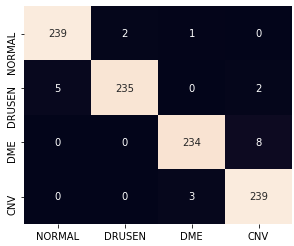}}
	\caption{Confusion Matrix for Classification with varying training samples.}\label{cm}
       \vspace{-0.4cm}
\end{figure}

The confusion matrix for Resnet classifier after being trained with 1\% stratified random samples and with the 2\% proposed sub-sampling method are shown in Fig. \ref{cm}.

Here, we observe that using the proposed sub-sampled data, some DME images get mis-classified as CNV where small cystic regions develop close to the inner sub-retinal layers. Also some images with thin drusen membranes are mistaken for normal images and some images with CNV and poor contrast are classified as DME. Most of these classifications can be improved by pre-processed zooming in and contrast corrections applied on test images in future works.

%%%% Update above.
\subsection{Explainability of Pathology}
\begin{figure}[ht]
    \centering
	\subfigure[Examples of proposed sub-sampled images along cluster peripheries.]
	{\includegraphics[height=1.5in, width=2in]{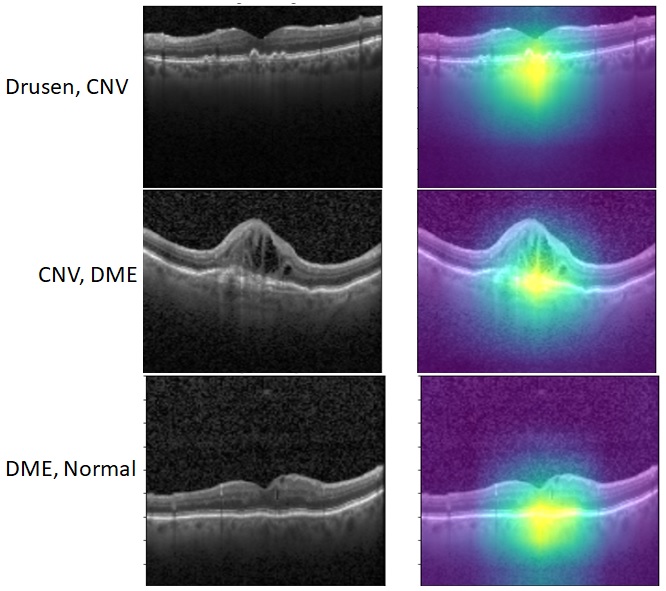}}
	\subfigure[ROI explanations for SimCLR augmentations shows the same ROI is highlighted per augmentation.]
	{\includegraphics[width=0.25\textwidth]{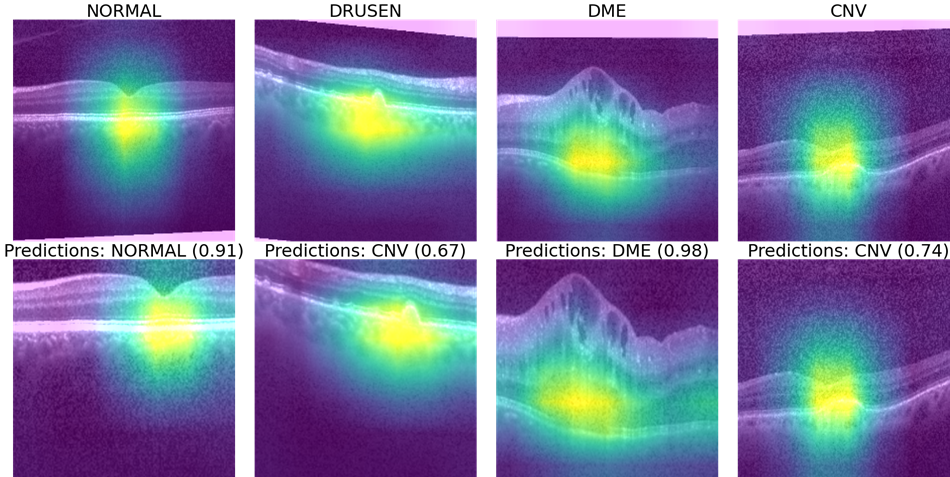}}
	\caption{Examples of pathology explanation on (a) sub-sampled training images, (b) test images, respectively.}\label{explain}
       \vspace{-0.4cm}
\end{figure}

Finally, we incorporate the Gradcam library to analyze the ROI that contributes to pathology classification. In Fig. \ref{explain} (a), we observe some sub-sampled images using Algorithm \ref{algo} that lie along the border of two distinct clusters, thereby enhancing classification performance for samples of those clusters. Further, Fig. \ref{explain} (b) shows the spatial robustness of the fine-tuned Resnet for pathology explainability.
\section{Conclusion}\label {conclusion}
In this work we present a novel semi-supervised algorithm that uses encoded latent features per image to identify a minimal sub-sampled data set that can be useful to train and visually explain pathology.
We incorporate a Resnet model for pathology classification that is first trained using contrastive loss to distinguish an original and augmented sample from other images. Next, we perform label spreading on the encoded latent space to identify a training set of upto 2\% of samples that represent the most uncertainty in feature space. Once trained on this labelled minimal saub-sampled data, the final classifier can achieve upto 97\% classification accuracy and is capable of visually explaining pathological sites. Future work can be directed towards extending the proposed image sub-sampling method and Resnet training to other pathologies.

\section*{Acknowledgement}
We gratefully acknowledge the direction setting and help by the AI team at Samsung SDSA.

\bibliographystyle{IEEEtran}
\bibliography{main}

\begin{thebibliography}{1}
\providecommand{\url}[1]{#1}
\csname url@samestyle\endcsname
\providecommand{\newblock}{\relax}
\providecommand{\bibinfo}[2]{#2}
\providecommand{\BIBentrySTDinterwordspacing}{\spaceskip=0pt\relax}
\providecommand{\BIBentryALTinterwordstretchfactor}{4}
\providecommand{\BIBentryALTinterwordspacing}{\spaceskip=\fontdimen2\font plus
\BIBentryALTinterwordstretchfactor\fontdimen3\font minus
  \fontdimen4\font\relax}
\providecommand{\BIBforeignlanguage}[2]{{%
\expandafter\ifx\csname l@#1\endcsname\relax
\typeout{** WARNING: IEEEtran.bst: No hyphenation pattern has been}%
\typeout{** loaded for the language `#1'. Using the pattern for}%
\typeout{** the default language instead.}%
\else
\language=\csname l@#1\endcsname
\fi
#2}}
\providecommand{\BIBdecl}{\relax}
\BIBdecl

\bibitem{kagglepaper}
D.~S. Kermany, M.~Goldbaum, W.~Cai, C.~C. Valentim, H.~Liang, S.~L. Baxter,
  A.~McKeown, G.~Yang, X.~Wu, F.~Yan \emph{et~al.}, ``Identifying medical
  diagnoses and treatable diseases by image-based deep learning,'' \emph{Cell},
  vol. 172, no.~5, pp. 1122--1131, 2018.

\bibitem{capsulenet}
T.~Tsuji, Y.~Hirose, K.~Fujimori, T.~Hirose, A.~Oyama, Y.~Saikawa, T.~Mimura,
  K.~Shiraishi, T.~Kobayashi, A.~Mizota \emph{et~al.}, ``Classification of
  optical coherence tomography images using a capsule network,'' \emph{BMC
  ophthalmology}, vol.~20, no.~1, pp. 1--9, 2020.

\bibitem{simclr}
T.~Chen, S.~Kornblith, M.~Norouzi, and G.~Hinton, ``A simple framework for
  contrastive learning of visual representations,'' 2020.

\bibitem{activelearning}
S.~Dasgupta and D.~Hsu, ``Hierarchical sampling for active learning,'' in
  \emph{Proceedings of the 25th international conference on Machine learning},
  2008, pp. 208--215.

\bibitem{hao2021}
H.~Hao, S.~Didari, J.~O. Woo, H.~Moon, and P.~Bangert, ``Highly efficient
  representation and active learning framework for imbalanced data and its
  application to covid-19 x-ray classification,''
  \emph{https://arxiv.org/abs/2103.05109}, 2021.

\bibitem{PLL}
K.~Sun, Z.~Min, and J.~Wang, ``Pp-pll: Probability propagation for partial
  label learning,'' in \emph{Joint European Conference on Machine Learning and
  Knowledge Discovery in Databases}.\hskip 1em plus 0.5em minus 0.4em\relax
  Springer, 2019, pp. 123--137.

\bibitem{Resnet}
C.~Szegedy, S.~Ioffe, V.~Vanhoucke, and A.~Alemi, ``Inception-v4,
  inception-resnet and the impact of residual connections on learning,'' in
  \emph{Proceedings of the AAAI Conference on Artificial Intelligence},
  vol.~31, no.~1, 2017.

\bibitem{gradcam}
\BIBentryALTinterwordspacing
R.~Meudec. (Accessed, Jan, 2021) tf\_explain. [Online]. Available:
  \url{https://github.com/sicara/tf-explain}
\BIBentrySTDinterwordspacing

\end{thebibliography}

\end{document}